\documentclass[runningheads]{llncs}
\usepackage{graphicx}
\usepackage{amsmath,amssymb} 
\usepackage{color}

\usepackage{times}
\usepackage{epsfig}

\usepackage{cite}
\usepackage{soul}
\usepackage{multirow}
\usepackage{rotating}
\usepackage{algorithm}
\usepackage[noend]{algpseudocode}
\usepackage{subfig}
\usepackage{enumitem}
\usepackage{subfig}
\setlist{nosep}

\usepackage{hyperref}

\ifx\mjdraft\undefined

\else

\fi

\def\eg{{e.g.}} 
\def\ie{{i.e.}}

\def\etal{\emph{et~al.\ }}
\newcommand{\SOTA}{state-of-the-art }
\newcommand{\myparagraph}[1]{\noindent \textbf{#1}}

\newcommand{\afterCaption}{\vspace{-0.3cm}}

\begin{document}
\pagestyle{headings}
\mainmatter
\def\ECCV18SubNumber{1875}  

\title{End-to-End Incremental Learning}
\titlerunning{End-to-End Incremental Learning}


\author{Francisco M. Castro \inst{1} \and
Manuel J. Mar\'{i}n-Jim\'{e}nez \inst{2}
\and
Nicol\'{a}s Guil \inst{1} \and
Cordelia Schmid \inst{3} \and
Karteek Alahari \inst{3}}

\authorrunning{Castro et al.}

\institute{
Department of Computer Architecture, University of M\'{a}laga, M\'{a}laga, Spain \\
\and
Department of Computing and Numerical Analysis, University of C\'{o}rdoba, C\'{o}rdoba, Spain\\
\and
Univ.\ Grenoble Alpes, Inria, CNRS, Grenoble INP, LJK, 38000 Grenoble, France\\
}

\maketitle

\begin{abstract}
   Although deep learning approaches have stood out in recent years due to their state-of-the-art results, they continue to suffer from {\it catastrophic forgetting}, a dramatic decrease in overall performance when training with new classes added incrementally. This is due to current neural network architectures requiring the entire dataset, consisting of all the samples from the old as well as the new classes, to update the model---a requirement that becomes easily unsustainable as the number of classes grows. We address this issue with our approach to learn deep neural networks incrementally, using new data and only a small exemplar set corresponding to samples from the old classes. This is based on a loss composed of a distillation measure to retain the knowledge acquired from the old classes, and a cross-entropy loss to learn the new classes. Our incremental training is achieved while keeping the entire framework end-to-end, i.e., learning the data representation and the classifier jointly, unlike recent methods with no such guarantees. We evaluate our method extensively on the CIFAR-100 and ImageNet (ILSVRC 2012) image classification datasets, and show state-of-the-art performance.
   
\end{abstract}

\section{Introduction}\label{sec:introduction}

One of the main challenges in developing a visual recognition system targeted at real-world applications is learning classifiers incrementally, where new classes are learned continually. For example, a face recognition system must handle new faces to identify new people. This task needs to be accomplished without having to re-learn faces already learned. While this is trivial to accomplish for most people (we learn to recognize faces of new people we meet every day), it is not the case for a machine learning system. Traditional models require all the samples (corresponding to the old and the new classes) to be available at training time, and are not equipped to consider only the new data, with a small selection of the old data. In an ideal system, the new classes should be integrated into the existing model, sharing the previously learned parameters. Although some attempts have been made to address this, most of the previous models still suffer from a dramatic decrease in performance on the old classes when new information is added, in particular, in the case of deep learning approaches~\cite{mccloskey1989catastrophic,goodfellow2013catastrophic,french1994dynamically,ans2004self,ratcliff1990connectionist,lopez2017nips,li2016eccv,rebuffi2017cvpr,shmelkov17iccv}. We address this challenging task in this paper using the problem of image classification to illustrate our results.

A truly incremental deep learning approach for classification is characterized by its: \textit{(i)}~ability to being trained from a flow of data, with classes appearing in any order, and at any time; \textit{(ii)}~good performance on classifying old and new classes; \textit{(iii)}~reasonable number of parameters and memory requirements for the model; and \textit{(iv)}~end-to-end learning mechanism to update the classifier and the feature representation jointly. Therefore, an ideal approach would be able to train on an infinitely-large number of classes in an incremental way, without losing accuracy, and having exactly the same number of parameters, as if it were trained from scratch.

None of the existing approaches for incremental learning~\cite{ruping2001icdm,cauwenberghs2000nips,li2016eccv,jung2016less,shmelkov17iccv,furlanello2016altm,Triki17iccv,rusu2016progressive,terekhov2015blocks,kirkpatrick2017pnas,xiao2014m,rebuffi2017cvpr} satisfy all these constraints. They often
decouple the classifier and representation learning tasks~\cite{rebuffi2017cvpr}, or are limited to very specific situations, \eg, learning from new datasets but not new classes related to the old ones~\cite{li2016eccv,jung2016less,furlanello2016altm,Triki17iccv}, or particular problems, \eg, object detection~\cite{shmelkov17iccv}. Some of them~\cite{ruping2001icdm, cauwenberghs2000nips} are tied to traditional classifiers such as SVMs and are unsuitable for deep learning architectures. Others~\cite{rusu2016progressive,terekhov2015blocks,kirkpatrick2017pnas,xiao2014m} lead to a rapid increase in the number of parameters or layers, resulting in a large memory footprint as the number of classes increases. In summary, there are no state-of-the-art methods that satisfy all the characteristics of a truly incremental learner.

The main contribution of this paper is addressing this challenge with our end-to-end approach designed specifically for incremental learning. The model can be realized with any deep learning architecture, together with our representative memory component, which is akin to an exemplar set for maintaining a small set of samples corresponding to the old classes (see Sec.~\ref{sec:memory}). The model is learned by minimizing the cross-distilled loss, a combination of two loss functions: cross-entropy to learn the new classes and distillation to retain the previous knowledge corresponding to the old classes (see Sec.~\ref{sec:deep_model}). As detailed in Sec.~\ref{sec:training}, any deep learning architecture can be adapted to our incremental learning framework, with the only requirement being the replacement of its original loss function with our new incremental loss. Finally, we illustrate the effectiveness of our image classification approach in obtaining \SOTA results for incremental learning on CIFAR-100~\cite{krizhevsky2009cifar} and ImageNet~\cite{russakovsky2015imagenet} (see Sec.~\ref{sec:experiments_cifar} and Sec.~\ref{sec:experiments_imagenet}).

\section{Related Work}\label{sec:relwork}
We now describe methods relevant to our approach by organizing them into traditional ones using a fixed feature set, and others that learn the features through deep learning frameworks, in addition to training classifiers.

\myparagraph{\bf Traditional approaches.} Initial methods for incremental learning targeted the SVM classifier~\cite{cortes1995svm}, exploiting its core components: support vectors and Karush-Kuhn-Tucker conditions. Some of these~\cite{ruping2001icdm} retain the support vectors, which encode the classifier learned on old data, to learn the new decision boundary together with new data. Cauwenberghs and Poggio~\cite{cauwenberghs2000nips} proposed an alternative to this by retaining the Karush-Kuhn-Tucker conditions on all the previously seen data (which corresponds to the old classes), while updating the solution according to the new data. While these early attempts showed some success, they are limited to a specific classifier and also do not extend to the current paradigm of learning representations and classifiers jointly.

Another relevant approach is learning concepts over time, in the form of lifelong~\cite{thrun1998ll} or never-ending~\cite{mitchell2015aaai,chen2013iccv,divvala2014cvpr} learning. Lifelong learning is akin to transferring knowledge acquired on old tasks to the new ones. Never-ending learning, on the other hand, focuses on continuously acquiring data to improve existing classifiers or to learn new ones. Methods in both these paradigms either require the entire training dataset, e.g.,~\cite{chen2013iccv}, or rely on a fixed representation, e.g.,~\cite{divvala2014cvpr}. Methods such as~\cite{mensink2013pami,restin2014cvpr,ruvolo2013icml} partially address these issues by learning classifiers without the complete training set, but are still limited due to a fixed or engineered data representation. This is achieved by: \textit{(i)} restricting the classifier or regression models, e.g., those that are linearly decomposable~\cite{ruvolo2013icml}, or \textit{(ii)} using a nearest mean classifier (NMC)~\cite{mensink2013pami}, or a random forest variant~\cite{restin2014cvpr}. Incremental learning is then performed by updating the bases or the per-class prototype, i.e., 
the average feature vector of the observed data, respectively.

Overall, the main drawback of all these methods is the lack of a task-specific data representation, which results in lower performance. Our proposed method addresses this issue with joint learning of features and classifiers.

\myparagraph{\bf Deep learning approaches.} This class of methods provides a natural way to learn task-specific features and classifiers jointly~\cite{ren2015nips,simonyan2014nips,bengio2013pami}. However, learning models incrementally in this paradigm results in \textit{catastrophic forgetting}, a phenomenon where the performance on the original (old) set of classes degrades dramatically~\cite{mccloskey1989catastrophic,goodfellow2013catastrophic,french1994dynamically,ans2004self,ratcliff1990connectionist,lopez2017nips,li2016eccv,rebuffi2017cvpr,shmelkov17iccv}. Initial attempts to overcome this issue were aimed at connectionist networks~\cite{mccloskey1989catastrophic,french1994dynamically,ans2004self}, and are thus inapplicable in the context of today's deep architectures for computer vision problems.

A more recent attempt to preserve the performance on the old tasks was presented in~\cite{li2016eccv} using distillation loss in combination with the standard cross-entropy loss. Distillation loss, which was originally proposed to transfer knowledge between different neural networks~\cite{hinton2014nips}, was adapted to maintain the responses of the network on the old tasks whilst updating it with new training samples~\cite{li2016eccv}. Although this approach reduced forgetting to some extent, in particular, in simplistic scenarios where the old and the new samples come from different datasets with little confusion between them, its performance is far from ideal. This is likely due to a weak knowledge representation of the old classes, and not augmenting it with an exemplar set, as done in our method. Works such as~\cite{rebuffi2017cvpr,Triki17iccv} demonstrated this weakness of~\cite{li2016eccv} showing significant errors in a sequential learning scenario, where samples from new classes are continuously added, and in particular when the new and the old samples are from related distributions---the challenging problem we consider in this paper. 

Other approaches using distillation loss, such as~\cite{jung2016less}, propose to freeze some of the layers corresponding to the original model, thereby limiting its adaptability to new data. Triki~\etal\cite{Triki17iccv} build on the method in~\cite{li2016eccv} using an autoencoder to retain the knowledge from old tasks, instead of the distillation loss. This method was also evaluated in a restrictive scenario, where the old and the new networks are trained on different datasets, similar to~\cite{li2016eccv}. Distillation loss was also adopted for learning object detectors incrementally~\cite{shmelkov17iccv}. Despite its success for object detection, the utility of this specific architecture for more general incremental learning scenarios we target here is unclear.

Alternative strategies to mitigate catastrophic forgetting include, increasing the number of layers in the network to learn features for the new classes~\cite{rusu2016progressive,terekhov2015blocks}, or slowing down the learning rate selectively through per-parameter regularization~\cite{kirkpatrick2017pnas}. Xiao~\etal\cite{xiao2014m} also follow a related scheme and grow their tree-structured model incrementally as new classes are observed. The main drawback of all these approaches is the rapid increase in the number of parameters, which grows with the total number of weights, tasks, and the new layers. In contrast, our proposed model results in minimal changes to the size of the original network, as explained in Sec.~\ref{sec:method}.

Rebuffi~\etal\cite{rebuffi2017cvpr} present iCaRL, an incremental learning approach where the tasks of learning the classifier and the data representation are decoupled. iCaRL uses a traditional NMC to classify test samples, i.e., it maintains an auxiliary set containing old and new data samples. The data representation model, which is a standard neural network, is updated as and when new samples are available, using a combination of distillation and classification losses~\cite{hinton2014nips,li2016eccv}. While our approach also uses a few samples from the old classes as exemplars in the representative memory component (cf.\ Sec.~\ref{sec:memory}), it overcomes the limitations of previous work by learning the classifier and the features jointly, in an end-to-end fashion. Furthermore, as shown in Sec.~\ref{sec:experiments_cifar} and Sec.~\ref{sec:experiments_imagenet}, our new model outperforms~\cite{rebuffi2017cvpr}.

\section{Our Model}\label{sec:method}
\begin{figure}[tb]
    \centering
    \includegraphics[width=0.98\textwidth]{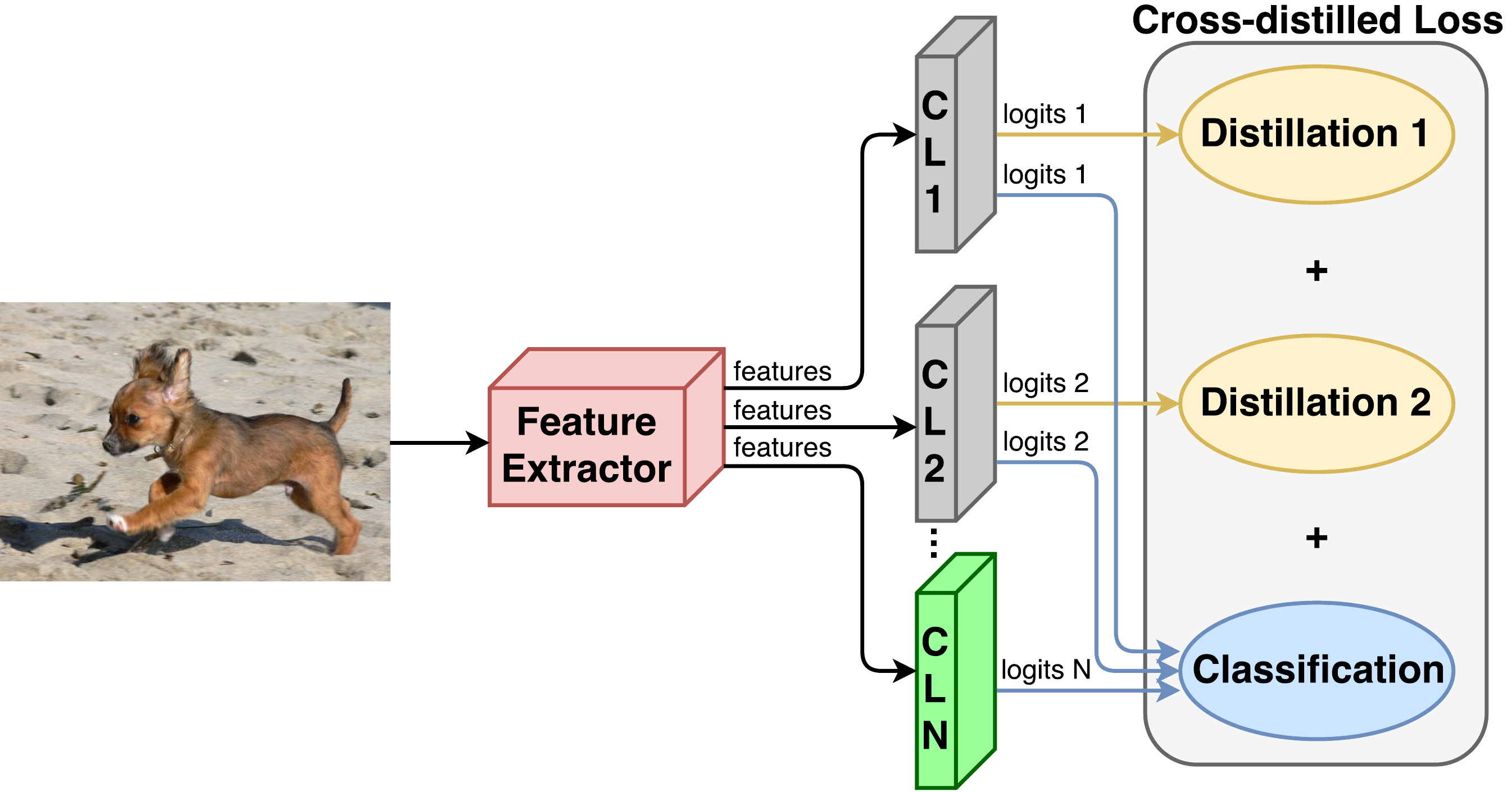}
    \caption{\textbf{Our incremental model}. Given an input image, the feature extractor produces a set of features which are used by the \textit{classification layers} (CL$i$ blocks) to generate a set of \textit{logits}. Grey \textit{classification layers} contain old classes and their \textit{logits} are used for distillation and classification. The green \textit{classification layer} (CL$N$ block) contains new classes and its \textit{logits} are involved only in classification. (Best viewed in color.)}
    \afterCaption
    \label{fig:model-loss}
\end{figure}

Our end-to-end approach uses a deep network trained with a cross-distilled loss function, i.e., cross-entropy together with distillation loss. The network can be based on the architecture of most deep models designed for classification, since our approach does not require any specific properties. A typical architecture for classification can be seen in Fig.~\ref{fig:model-loss}, with one \textit{classification layer} and a classification loss. This \textit{classification layer} uses features from the feature extractor to produce a set of \textit{logits} which are transformed into class scores by a softmax layer (not shown in the figure). The only necessary modification is the loss function, described in Sec.~\ref{sec:deep_model}. To help our model retain the knowledge acquired from the old classes, we use a representative memory (Sec.~\ref{sec:memory}) that stores and manages the most representative samples from the old classes. In addition to this we perform data augmentation and a balanced fine-tuning (Sec.~\ref{sec:training}). All these components put together allow us to get state-of-the-art results.

\subsection{Representative memory}\label{sec:memory}
When a new class or set of classes is added to the current model, a subset with the most representative samples from them is selected and stored in the representative memory. We investigate two memory setups in this work. The first setup considers a memory with a limited capacity of \textit{K} samples. As the capacity of the memory is independent of the number of classes, the more classes stored, the fewer samples retained per class. The number of samples per class, $n$, is thus given by $n = \lfloor K / c\rfloor$, where $c$ is the number of classes stored in memory, and $K$ is the memory capacity. The second setup stores a constant number of exemplars per class. Thus, the size of the memory grows with the number of classes.

The representative memory unit performs two operations: selection of new samples to store, and removal of leftover samples.

\myparagraph{Selection of new samples}. This is based on \textit{herding selection}~\cite{welling2009icml}, which produces a sorted list of samples of one class based on the distance to the mean sample of that class. Given the sorted list of samples, the first $n$ samples of the list are selected. These samples are most representative of the class according to the mean. This selection method was chosen based on our experiments testing different approaches, such as random selection, histogram of the distances from each sample to the class mean, as shown in Sec.~\ref{sec:experiments_val}. The selection is performed once per class, whenever a new class is added to the memory.

\myparagraph{Removing samples}. 
This step is performed after the training process to allocate memory for the samples from the new classes. As the samples are stored in a sorted list, this operation is trivial. The memory unit only needs to remove samples from the end of the sample set of each class. Note that after this operation, the removed samples are never used again.

\subsection{Deep network}\label{sec:deep_model}
\myparagraph{Architecture.} The network is composed of several components, as illustrated in Fig.~\ref{fig:model-loss}. The first component is a \textit{feature extractor}, which is a set of layers to transform the input image into a feature vector. The next component is a \textit{classification layer} which is the last fully-connected layer of the model, with as many outputs as the number of classes. This component takes the features and produces a set of \textit{logits}. During the training phase, gradients to update the weights of the network are computed with these \textit{logits} through our cross-distilled loss function. At test time, the loss function is replaced by a softmax layer (not shown in the figure).

To build our incremental learning framework, we start with a traditional, \ie, non-incremental, deep architecture for classification for the first set of classes. When new classes are trained, we add a new \textit{classification layer} corresponding to these classes, and connect it to the \textit{feature extractor} and the component for computing the cross-distilled loss, as shown in Fig.~\ref{fig:model-loss}. Note that the architecture of the \textit{feature extractor} does not change during the incremental training process, and only new \textit{classification layers} are connected to it. Therefore, any architecture (or even pre-trained model) can be used with our approach just by adding the incremental classification layers and the cross-distilled loss function when necessary.

\myparagraph{Cross-distilled loss function.} This combines a distillation loss~\cite{hinton2014nips}, which retains the knowledge from old classes, with a multi-class cross-entropy loss, which learns to classify the new classes. The distillation loss is applied to the \textit{classification layers} of the old classes while the multi-class cross-entropy is used on all \textit{classification layers}. This allows the model to update the decision boundaries of the classes. The loss computation is illustrated in Fig.~\ref{fig:model-loss}. The cross-distilled loss function $L(\omega)$ is defined as:
\begin{equation}
    L(\omega) = L_{C}(\omega) + \sum_{f=1}^{F} L_{{D}_{f}}(\omega),
\label{eq:loss_function}
\end{equation}
where $L_{C}(\omega)$ is the cross-entropy loss applied to samples from the old and new classes, $L_{{D}_{f}}$ is the distillation loss of the \textit{classification layer} $f$, and $F$ is the total number of \textit{classification layers} for the old classes (shown as grey boxes in Fig.~\ref{fig:model-loss}). 

The cross-entropy loss $L_{C}(\omega)$ is given by:
\begin{equation}\label{eq:cross-entropy}
    L_{C}(\omega) = -\frac{1}{N} \sum_{i=1}^{N} \sum_{j=1}^{C} p_{ij} \log q_{ij},
\end{equation}
where $q_i$ is a score obtained by applying a softmax function to the \textit{logits} of a \textit{classification layer} for sample $i$, $p_i$ is the ground truth for the sample $i$, and $N$ and $C$ denote the number of samples and classes respectively.

The distillation loss $L_{D}(\omega)$ is defined as:
\begin{equation}
    L_{D}(\omega) = -\frac{1}{N} \sum_{i=1}^{N} \sum_{j=1}^{C} pdist_{ij} \log qdist_{ij},
\label{eq:distillation_loss}
\end{equation}
where $pdist_i$ and $qdist_i$ are modified versions of $p_i$ and $q_i$, respectively. They are obtained by raising $p_i$ and $q_i$ to the exponent $1/T$, as described in~\cite{hinton2014nips}, where $T$ is the distillation parameter. When $T=1$, the class with the highest score influences the loss significantly, e.g., more than $0.9$ from a maximum of $1.0$, and the remaining classes with low scores have minimal impact on the loss. However, with $T>1$, the remaining classes have a greater influence, and their higher loss values must be minimized. This forces the network to learn a more fine grained separation between them. As a result, the network learns a more discriminative representation of the classes. Based on our empirical results, we set $T$ to 2 for all our experiments.

\section{Incremental Learning}\label{sec:training}
\begin{figure}[tb]
    \centering
    \includegraphics[width=0.98\textwidth]{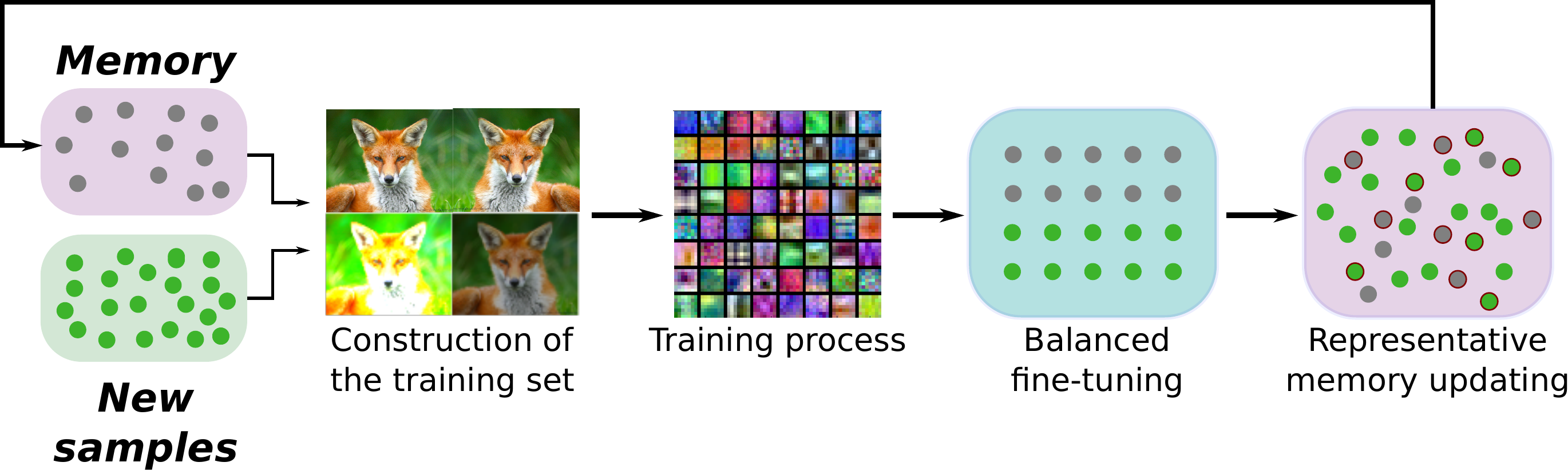}
    \caption{\textbf{Incremental training}. Grey dots correspond to samples stored in the representative memory. Green dots correspond to samples from the new classes. Dots with red border correspond to the selected samples to be stored in the memory. (Best viewed in color.)}
    \afterCaption
    \label{fig:training}
\end{figure}

An incremental learning step in our approach consists of four main stages, as illustrated in Fig.~\ref{fig:training}. The first stage is the construction of the training set, which prepares the training data to be used in the second stage, the training process, which fits a model given the training data. In the third stage, a fine-tuning with a subset of the training data is performed. This subset contains the same number of samples per class. Finally, in the fourth stage, the representative memory is updated to include samples from the new classes. We now describe these stages in detail.

\myparagraph{Construction of the training set.} Our training set is composed of samples from the new classes and exemplars from the old classes stored in the representative memory. As our approach uses two loss functions, \ie, classification and distillation, we need two labels for each sample, associated with the two losses. For classification, we use the one-hot vector which indicates the class appearing in the image. For distillation, we use as labels the \textit{logits} produced by every \textit{classification layer} with old classes (grey fully-connected layers in Fig.~\ref{fig:model-loss}). Thus, we have as many distillation labels per sample as \textit{classification layers} with old classes. To reinforce the old knowledge, samples from the new classes are also used for distillation. This way, all images produce gradients for both the losses. Thus, when an image is evaluated by the network, the output encodes the behaviour of the weights that compose every layer of the deep model, independently of its label. Each image of our training set will have a classification label and $F$ distillation labels; cf.\ Eq.~\ref{eq:loss_function}. Note that this label extraction is performed in each incremental step.

Consider an example scenario to better understand this step, where we are performing the third incremental step of our model (Fig.~\ref{fig:model-loss}). At this point the model has three \textit{classification layers} ($N=3$), two of them will process old classes (grey boxes), i.e., $F=2$, and one of them operates on the new classes (green box). When a sample is evaluated, the \textit{logits} produced by the two \textit{classification layers} with the old classes are used for distillation (yellow arrows), and the \textit{logits} produced by the three \textit{classification layers} are used for classification (blue arrows).

\myparagraph{Training process.} Our cross-distilled loss function (Eq.~\ref{eq:loss_function}) takes the augmented training set with its corresponding labels and produces a set of gradients to optimise the deep model. Note that, during training, all the weights of the model are updated. Thus, for any sample, features obtained from the feature extractor are likely to change between successive incremental steps, and the \textit{classification layers} should adapt their weights to deal with these new features. This is an important difference with some other incremental approaches like~\cite{li2016eccv}, where the the \textit{feature extractor} is frozen and only the \textit{classification layers} are trained.

\myparagraph{Balanced fine-tuning.} Since we do not store all the samples from the old classes, samples from these classes available for training can be significantly lower than those from the new classes. To deal with this unbalanced training scenario, we add an additional fine-tuning stage with a small learning rate and a balanced subset of samples. The new training subset contains the same number of samples per class, regardless of whether they belong to new or old classes. This subset is built by reducing the number of samples from the new classes, keeping only the most representative samples from each class, according to the selection algorithm described in Sec.~\ref{sec:memory}. With this removal of samples from the new classes, the model can potentially forget knowledge acquired during the previous training step. We avoid this by adding a temporary distillation loss to the \textit{classification layer} of the new classes.

\myparagraph{Representative memory updating.} After the balanced fine-tuning step, the representative memory must be updated to include exemplars from the new classes. This is performed with the selection and removing operations described in Sec.~\ref{sec:memory}. First, the memory unit removes samples from the stored classes to allocate space for samples from the new classes. Then, the most representative samples from the new classes are selected, and stored in the memory unit according to the selection algorithm.

\section{Implementation Details}\label{sec:implementation}
Our models are implemented on MatConvNet~\cite{vedaldi2015matconvnet}. For each incremental step, we perform 40 epochs, and an additional 30 epochs for balanced fine-tuning. Our learning rate for the first 40 epochs starts at $0.1$, and is divided by $10$ every $10$ epochs. The same reduction is used in the case of fine-tuning, except that the starting rate is $0.01$. We train the networks using standard stochastic gradient descent with mini-batches of $128$ samples, weight decay of $0.0001$ and momentum of $0.9$. We apply $L^{2}$-regularization and random noise~\cite{neelakantan2017noise} (with parameters $\eta = 0.3, \gamma = 0.55$) on the gradients to minimize overfitting.

Following the setting suggested by He~\etal\cite{he2015resnet}, we use dataset-specific CNN/deep models. This allows the architecture of the network to be adapted to specific characteristics of the dataset. We use a $32$-layer ResNet for CIFAR-100, and a $18$-layer ResNet for ImageNet as the deep model. We store $K = 2000$ distillation samples in the representative memory for CIFAR-100 and $K = 20000$ for ImageNet. When training the model for CIFAR-100, we normalize the input data by dividing the pixel values by $255$, and subtracting the mean image of the training set. In the case of ImageNet, we only perform the subtraction, without the pixel value normalization, following the implementation of~\cite{he2015resnet}.

Since there are no readily-available class-incremental learning benchmarks, we follow the standard setup~\cite{rebuffi2017cvpr,shmelkov17iccv} of splitting the classes of a traditional multi-class dataset into incremental batches. In all the experiments below, iCaRL refers to the final method in~\cite{rebuffi2017cvpr}, and hybrid1 refers to their variant, which uses a CNN classifier instead of NMC. LwF.MC is the multi-class implementation of LwF~\cite{li2016eccv}, as done in~\cite{rebuffi2017cvpr}. We used the publicly available implementation of iCaRL from GitHub\footnote{\label{footnote:code}\url{https://github.com/srebuffi/iCaRL}}. The results for LwF.MC are also obtained from this code, without the exemplar usage. We report results for each method as the average accuracy over all the incremental batches. Note that we do not consider the accuracy of the first batch in this average, as it does not correspond to incremental learning. This is unlike the evaluation in~\cite{rebuffi2017cvpr}, which is the reason for difference between the results we report for their method, and the published results.

\myparagraph{Data augmentation.} The second and third stages of our approach (cf.\ Sec.~\ref{sec:training}) perform data augmentation before the training step. Specifically, the operations performed are:
\begin{enumerate}
    \item \textit{Brightness}: the intensity of the original image is altered by adding a random intensity value in the range $[-63, 63]$.
    \item \textit{Contrast normalization}: the contrast of the original image is altered by a random value in the range $[0.2, 1.8]$. The operation performed is $im_{\mathrm{altered}} = (im  - mean) \times contrast + mean$. Where $im$ is the original image, $mean$ is the mean of the pixels per channel, and $contrast$ is the random contrast value.
    \item \textit{Random cropping}: all the images (original, brightness and contrast) are randomly cropped.
    \item \textit{Mirroring}: a mirror image is computed for all images (original, brightness, contrast and crops).
\end{enumerate}
Other operations applied on each dataset are specified in Sec.~\ref{sec:experiments_cifar} for CIFAR-100 and in Sec.~\ref{sec:experiments_imagenet} for ImageNet.
\section{Evaluation on CIFAR-100} \label{sec:experiments_cifar}
We perform three types of experiments on the CIFAR-100 dataset. In the first one (Sec.~\ref{sec:results_cifar}), we set the maximum storage capacity of our representative memory unit, following the experimental protocol in~\cite{rebuffi2017cvpr}. The second experiment (Sec.~\ref{sec:incremental_analysis}) evaluates the methods without a fixed memory size, and uses a constant number samples for each of the old classes instead. Here, the memory size grows with each incremental step, when new classes are stored in the representative memory unit. Finally, in Sec.~\ref{sec:experiments_val}, we perform an ablation study to analyze the influence of different components of our approach on the accuracy.

\myparagraph{Dataset.}
CIFAR-100 dataset~\cite{krizhevsky2009cifar} is composed of $60$k $32\times32$ RGB images of $100$ classes, with $600$ images per class. Every class has $500$ images for training and $100$ images for testing. We divide the $100$ classes into splits of $2$, $5$, $10$, $20$, and $50$ classes with a random order. Thus, we will have $50$, $20$, $10$, $5$, and $2$ incremental training steps respectively. After each incremental step, the resulting model is evaluated on the test data composed of all the trained classes, \ie, old and new ones. Our evaluation metric at each incremental step is the standard multi-class accuracy. We execute the experiments five times with different random class orders, reporting the average accuracy and standard deviation. In addition, we report the average incremental accuracy (mean of the accuracy values at every incremental step). As mentioned earlier, we do not consider the accuracy of the first step for this average as it does not represent incremental learning.

On CIFAR, we follow the data augmentation steps described in Sec.~\ref{sec:implementation} and, for each training sample, generate 11 new samples: one brightness normalization, one contrast normalization, three random crops (applied to the original, brightness and contrast images) and six mirrors (applied to the previously generated images and the original one). 

\subsection{Fixed memory size} \label{sec:results_cifar}
We evaluate five different splits with different class order and incremental steps of $2$, $5$, $10$, $20$, and $50$ classes. The class order is identical for all the evaluated methods, to ensure that the results are comparable. Tab.~\ref{tab:SOTA}(a) summarises the results of the experiments and Fig.~\ref{fig:results} shows the incremental steps for $2$ and $5$ classes. The rest of plots are included in the appendix~\cite{suppmat}. The `Upper-Bound' result, shown in Fig.~\ref{fig:results} with a large cross (in magenta) in the last step, is obtained by training a non-incremental model using all the classes, and all their training samples.

We observe that our end-to-end approach obtains the best results for $2$, $5$, $10$, and $20$ classes. For $50$ classes, although we achieve the same score as Hybrid1 (the variant of iCaRL using CNN classifier), we are $1\%$ lower than iCaRL. This behaviour is due to the limited memory size, resulting in a heavily unbalanced training set containing $12.5$ times more data from the new samples than from the old classes. To highlight the statistical significance of our method's performance compared to iCaRL, we performed a paired $t$-test on the results obtained for CIFAR-100. The corresponding $p$-values are $0.00005$, $0.0005$, $0.003$, $0.0077$, $0.9886$ for $2$, $5$, $10$, $20$, and $50$ classes respectively, which shows that the improvement of our method over iCaRL is statistically significant ($p < 0.01$) in all cases, except for $50$ classes where both methods show similar performance.

\begin{figure*}[t]
    \centering
    \includegraphics[width=1.0\textwidth]{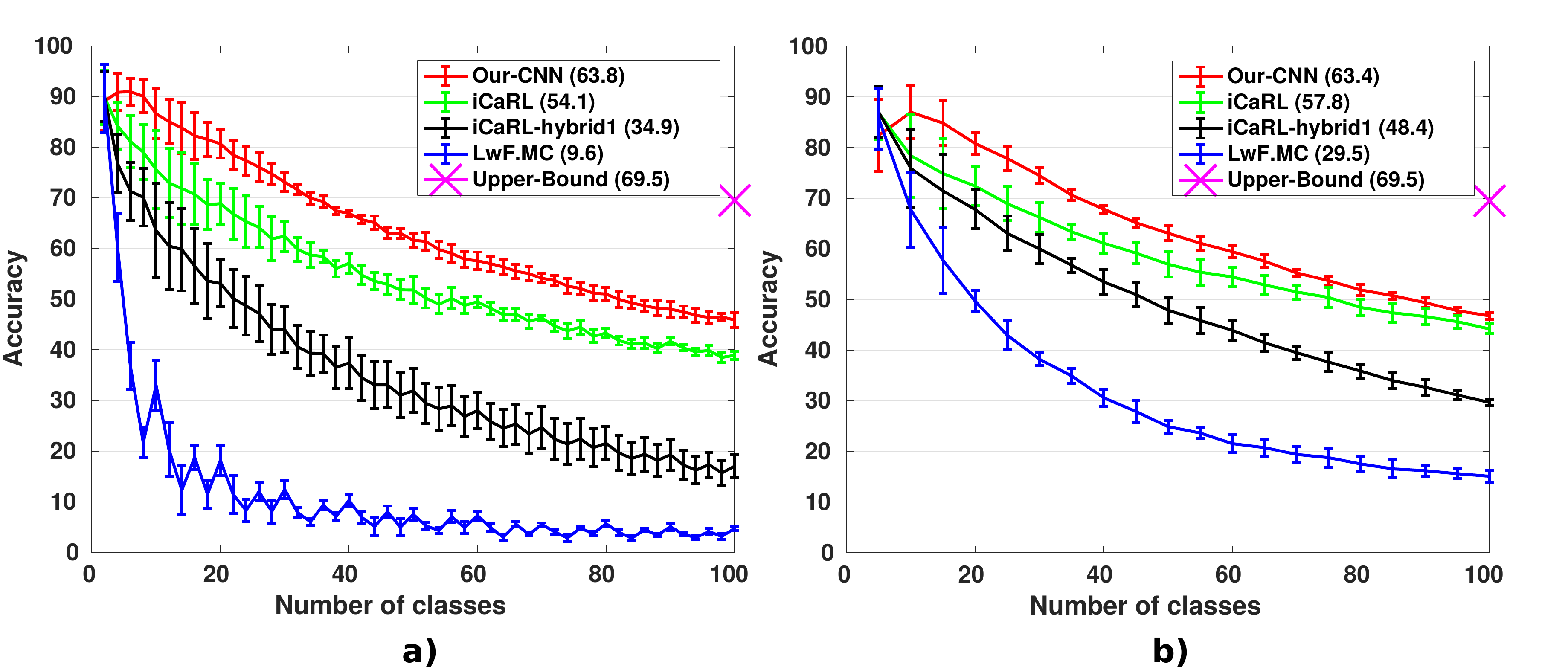}
    \caption{\textbf{Accuracy on CIFAR-100.} Average and standard deviation of 5 executions with (a) 2 and (b) 5 classes per incremental step. Average of the incremental steps is shown in parentheses for each method. (Best viewed in pdf.)}
    \afterCaption
    \label{fig:results}
\end{figure*}
\begin{table}[t]
\footnotesize
\centering
\setlength{\tabcolsep}{0.2em} %
\subfloat[][{CIFAR-100}]{
\begin{tabular}[t]{l|c|c|c|c|c}
\hline 
\# classes & 2 & 5 & 10 & 20 & 50 \\ 
\hline \hline
Our-CNN & \textbf{63.8 $\pm$ 1.9} & \textbf{63.4 $\pm$ 1.6} & \textbf{63.6 $\pm$ 1.3} & \textbf{63.7 $\pm$ 1.1} & 60.8 $\pm$ 0.3 \\
iCaRL & 54.1 $\pm$ 2.5 & 57.8 $\pm$ 2.6 & 60.5 $\pm$ 1.6 & 62.0 $\pm$ 1.2 & \textbf{61.8 $\pm$ 0.4} \\ 
Hybrid1 & 34.9 $\pm$ 4.5 & 48.4 $\pm$ 2.6 & 55.8 $\pm$ 1.8 & 60.4 $\pm$ 1.0 & 60.8 $\pm$ 0.7 \\ 
LwF.MC & 9.6 $\pm$ 1.5 & 29.5 $\pm$ 2.2 & 40.4 $\pm$ 2.0 & 47.6 $\pm$ 1.5 & 52.9 $\pm$ 0.6 \\ 
\hline 
\end{tabular}}
\quad
\subfloat[][{ImageNet}]{
\begin{tabular}[t]{l|c|c}
\hline
\# classes & 10 & 100 \\ 
\hline \hline
Our-CNN & \textbf{90.4} & \textbf{69.4} \\
iCaRL & 85.0 & 62.5 \\
Hybrid1 & 83.5 & 46.1 \\
LwF.MC & 79.1 & 43.8 \\
\hline 
\end{tabular}}
\caption{\textbf{Fixed memory size: accuracy on CIFAR-100 and ImageNet.} Each column represents a different number of classes per incremental step. Each row represents a different approach. The best results are marked in bold.}
\afterCaption
\afterCaption
\afterCaption
\label{tab:SOTA}
\end{table}

It can be also observed that the performance of our approach remains stable across the incremental step sizes (from 2 to 20 classes per step) in Tab.~\ref{tab:SOTA}(a), in contrast to all the other methods, which are dependent on the number of classes added in each step. This is because a small number of classes at each incremental step benefits the accuracy in the early stages of the incremental learning process, as only a few classes must be classified. However, as more steps are applied to train all the classes, the accuracy of the final stages decreases.

The behaviour is reversed when larger number of classes are added in each incremental step. Lower accuracy values are seen during the early stages, but this is compensated with better values in the final stages. These effects can be seen in Fig.~\ref{fig:results}, where two different number of classes per incremental step (2 and 5) are visualized. Fig.~\ref{fig:results} also shows that our approach is significantly better than iCaRL when a small number of classes per incremental step are employed. With larger number of classes in each step, iCaRL approaches our performance, but still remains lower overall. Our approach clearly outperforms LwF.MC in all the cases, thus highlighting the importance of the representative memory in our model.

\subsection{Fixed number of samples} \label{sec:incremental_analysis}
In this experiment, we train the models using a constant number of training samples per old class. This limitation is not applied to the samples from the new classes. Thus, we allow the memory to grow in proportion to the number of classes, in contrast to the fixed memory experiment, where the memory size remains constant. Additionally, to measure the impact of the number of samples in the accuracy, we evaluate different number of samples per class: $50$, $75$ and $100$. We focus on experiments with incremental step values of $5$, $10$ and $20$ classes. We consider the same class order for both iCaRL and our approach to ensure that the results are comparable. We focus the comparison on iCaRL and Hybrid1 in this experiment, as LwF.MC shows a lower performance than these two methods; see Sec.~\ref{sec:results_cifar}.

Tab.~\ref{tab:SOTA-images}(a) summarizes the results of these experiments. The number of classes per incremental step is indicated in the first row of the table. The second row contains the number of exemplars per old class used during training. The remaining rows show the results of the different approaches evaluated. Comparing the results between Our-CNN and the methods developed in~\cite{rebuffi2017cvpr}, we see that in all scenarios our approach performs better. As in the `fixed memory size' experiment (Sec.~\ref{sec:results_cifar}), our approach achieves a similar average accuracy for incremental step sizes ranging from 5 to 20 classes, e.g., $62.4$, $62.7$, $63.3$ with 50 exemplars per class, showing its stability. To measure the impact of the number of exemplars per class on the training, we compare the results in Tab.~\ref{tab:SOTA}(a) with those in Tab.~\ref{tab:SOTA-images}(a). In all cases, the more the exemplars used for training, the better the accuracy obtained. For 50 exemplars, the results are worse than those in Tab.~\ref{tab:SOTA} because in the early incremental steps, the number of exemplars available is lower, and these initial models are under trained. This causes a chain effect, and the model obtained in the final stage is worse than expected, even when more exemplars are available.

\begin{table}[t]
\small
\centering
\subfloat[][{Fixed number of samples}]{
\begin{tabular}[t]{l|c|c|c|c|c|c|c|c|c}
\hline 
\# classes & \multicolumn{3}{c|}{5} & \multicolumn{3}{c|}{10} & \multicolumn{3}{c}{20} \\ 
\hline 
\# img / cls & 50 & 75 & 100 & 50 & 75 & 100 & 50 & 75 & 100 \\ 
\hline \hline
Our-CNN & \textbf{62.4} & \textbf{66.9} & \textbf{68.6} & \textbf{62.7} & \textbf{65.7} & \textbf{68.5} & \textbf{63.3} & \textbf{65.4} & \textbf{67.3} \\ 
iCaRL & 56.5 & 59.9 & 62.2 & 60.0 & 62.3 & 63.7 & 61.9 & 63.0 & 64.0 \\ 
Hybrid1 & 45.7 & 49.2 & 50.9 & 55.3 & 56.5 & 57.4 & 60.4 & 61.5 & 62.2 \\ 
\hline 
\end{tabular}}
\qquad 
\subfloat[][{Ablation study}]{
\begin{tabular}[t]{l|c|c|c}
\hline 
\# classes & 5 & 10 & 20 \\ 
\hline \hline
Our-CNN-Base & 57.0 & 53.7 & 50.1 \\ 
Our-CNN-DA & 59.2 & 57.9 & 57.2 \\ 
Our-CNN-BF & 57.9 & 58.1 & 57.1 \\ 
Our-CNN-Full & \textbf{63.8} & \textbf{64.0} & \textbf{63.2} \\
iCaRL & 58.8 & 60.9 & 61.2 \\ 
Hybrid1 & 48.7 & 55.1 & 59.8 \\ 
\hline 
\end{tabular}}
\caption{\textbf{Accuracy on CIFAR-100.} Each row represents a different approach. The best results are marked in bold. See the main text for more details.}
\afterCaption
\afterCaption
\afterCaption
\label{tab:SOTA-images}
\end{table}

\subsection{Ablation studies} \label{sec:experiments_val}
We now analyze the components of our approach and demonstrate their impact on the final performance. All these ablation studies are performed with the fixed memory setup. We first evaluate the sample selection strategy with an experiment using incremental steps of 10 classes and three methods for selecting samples: herding, random and histogram selection. Herding is our selection method, presented in Sec.~\ref{sec:memory}. Random selection refers to choosing samples to be stored in memory randomly. In the histogram selection strategy, samples are chosen according to their distance to the mean of their class. We first compute a histogram of distances, with ten bins, and assign each sample to one of these bins. We then select samples from each bin according to the proportion of samples it contains. From the results (herding: $63.6\%$, random: $63.1\%$, and histogram: $59.1\%$), herding and random selection strategies show the best performance.

\begin{figure}[tb]
    \centering
    \includegraphics[width=1.0\textwidth]{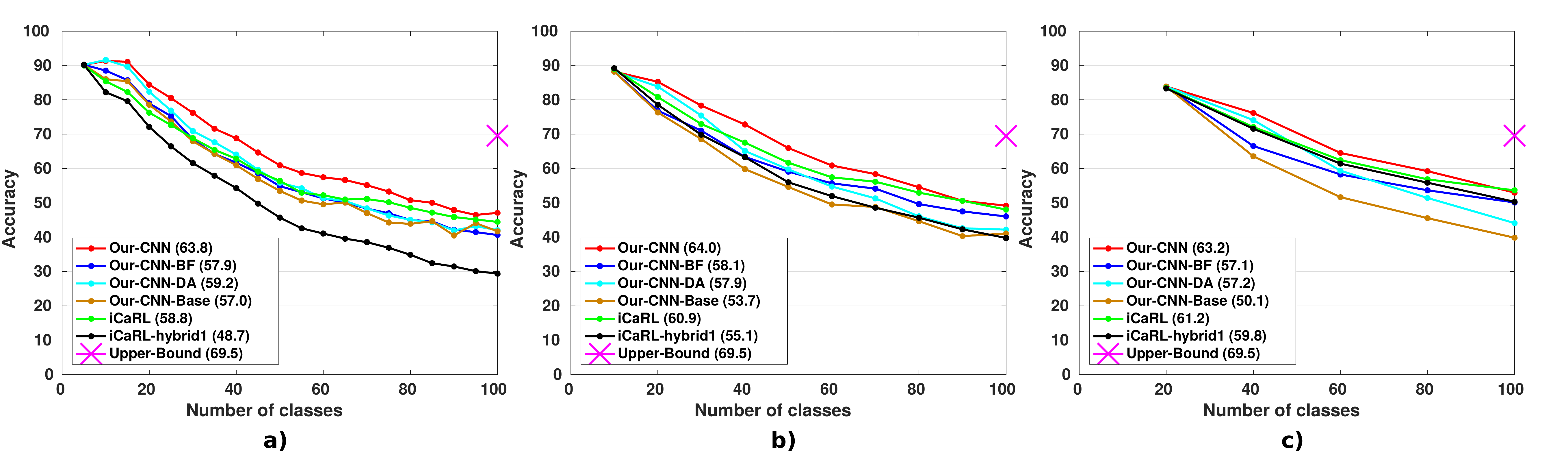}
    \caption{\textbf{Ablation study with CIFAR-100.} Results for {(a)} 5, {(b)} 10, and {(c)} 20 classes. The average over all the incremental steps is shown in parentheses for each method. (Best viewed in pdf.)}
    \afterCaption
    \label{fig:results-sup-ablation}
\end{figure}

In the following ablation study, we analyze the impact of augmentation and fine-tuning. We first train our model with data augmentation, but without balanced fine-tuning (`Our-CNN-DA'). In the second experiment, we train without data augmentation, but with balanced fine-tuning (`Our-CNN-BF'). Finally, we train our model without data augmentation and balanced fine-tuning (`Our-CNN-Base'). Here, we focus on experiments with incremental step values of $5$, $10$ and $20$ classes. As in previous experiments, the first split for iCaRL and our approach is run with the same order of classes, to ensure that the results are comparable. Tab.~\ref{tab:SOTA-images}(b) and Fig.~\ref{fig:results-sup-ablation} summarise the results for this study. The baseline `Our-CNN-Base' is the worst one for all cases. However, when the data augmentation (`Our-CNN-DA') is added, the results improve in all cases, obtaining the best result for 5 classes ($59.2$). However, due to the unbalanced number of samples between the old and new classes, with larger incremental steps it is necessary to add our balanced fine-tuning (`Our-CNN-BF'). When balanced fine-tuning (`Our-CNN-BF') is added, it improves the results in all cases, specially with big incremental steps, which highlights the importance of a balanced training set. Finally, when both the components are added to the baseline, obtaining our full model (`Our-CNN-Full'), we observe the best results and a new \SOTA is established on this dataset for incremental learning. 

\section{Evaluation on ImageNet} \label{sec:experiments_imagenet}
\myparagraph{Dataset.}
ImageNet Large-Scale Visual Recognition Challenge 2012 (ILSVRC12)~\cite{russakovsky2015imagenet} is an annual competition which uses a subset of ImageNet. This subset is composed of $1000$ classes with more than $1000$ images per class. In total, there are roughly 1.2 million training images, 50k validation images, and 150k testing images. We run two experiments with this dataset. In the first one, we randomly select $100$ classes, and divide them into splits of $10$ classes selected randomly. In the second one, we divide the $1000$ classes into splits of $100$ classes randomly selected. Note that the same set of classes are considered for all the approaches to ensure that the results are comparable. After every incremental step, the resulting model is evaluated on test data composed of all the trained classes. We execute the experiments once and report the top-5 accuracy for each incremental step. We also report the average incremental accuracy described in Sec.~\ref{sec:experiments_cifar}.

We use data augmentation described in Sec.~\ref{sec:implementation}, and for each training sample, generate its mirror image, and then randomly apply transformations (cf.\ Sec.~\ref{sec:implementation}) for all the images (original and mirror). Thus, with our data augmentation, we double the number of training samples.

\begin{figure*}[t]
    \centering
    \includegraphics[width=1.0\textwidth]{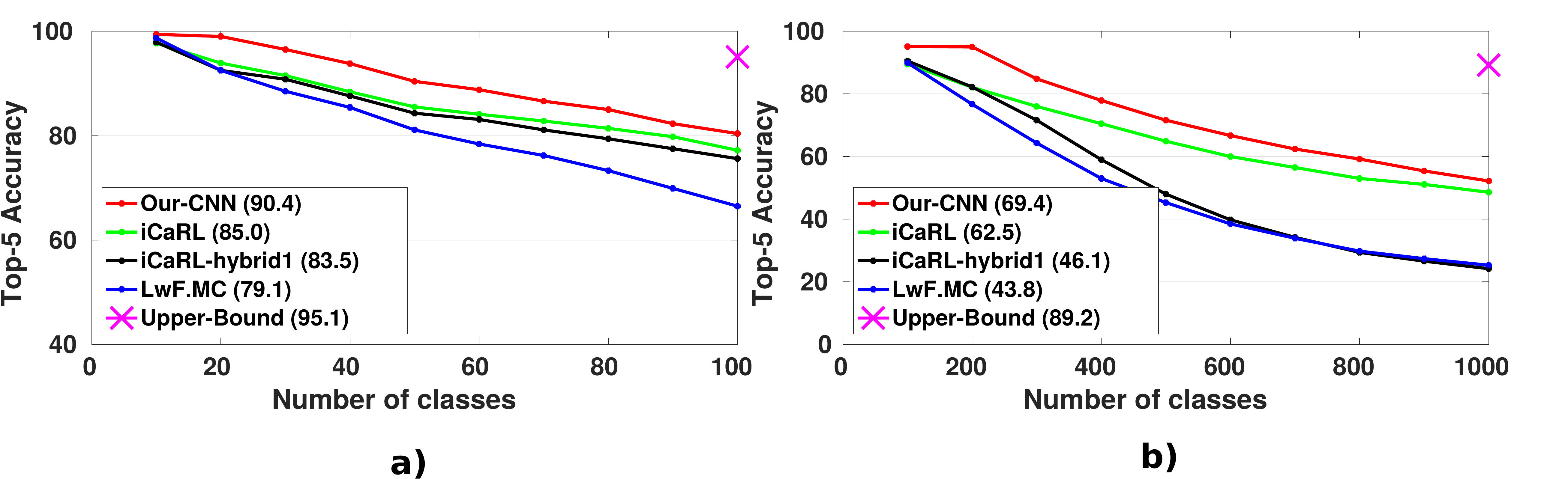}
    \caption{\textbf{Accuracy on ImageNet.} One execution with (a) 10 and (b) 100 classes per incremental step. Average of the incremental steps is shown in parentheses for each method. (Best viewed in pdf.)}
    \afterCaption
    \label{fig:results-imagenet}
\end{figure*}

\myparagraph{Fixed memory size.}
We maintain identical class order for all the evaluated methods, to ensure that the results are comparable. We also follow the protocol in~\cite{rebuffi2017cvpr} for a fair comparison with iCaRL and hybrid1. Tab.~\ref{tab:SOTA}(b) summarizes the results of this fixed memory size experiment, and Fig.~\ref{fig:results-imagenet} shows the incremental steps with $10$ and $100$ classes. The `Upper-Bound' result, shown with a cross in the figure, is obtained by training a non-incremental model using the training samples for all the classes. From the results, we observe that in both cases we establish a new \SOTA, improving the previous average results by more than $5\%$. This suggests that our approach is also suitable for large datasets with many classes. In addition, as the number of samples from new and old classes is more balanced than in CIFAR-100, our approach achieves good accuracy even with large incremental steps of $100$ classes.

\section{Summary}\label{sec:conclusion}
This paper presents a novel approach for training CNNs in an incremental fashion using a combination of cross-entropy and distillation loss functions. Experimental results on CIFAR-100 and ImageNet presented in the paper lead to the following conclusions. \textit{(i)} Our end-to-end approach is more robust than other recent methods, such as iCaRL, relying on a sub-optimal, independently-learned external classifier. \textit{(ii)} Representative memory, its size, and unbalanced training sets play an important role in the final accuracy. As part of future work we plan to explore new sample selection strategies, using a dynamic number of samples per class.

\paragraph{Acknowledgements.}
This work was supported in part by the projects TIC-1692 (Junta de Andaluc\'ia), TIN2016-80920R (Spanish Ministry of Science and Tech.), ERC advanced grant ALLEGRO, and EVEREST (no.\ 5302-1) funded by CEFIPRA. We gratefully acknowledge the support of NVIDIA Corporation with the donation of a Titan X Pascal GPU used for this research.

\bibliographystyle{splncs04}
\bibliography{references}

\clearpage
\section*{Appendix A: Additional results on CIFAR-100}

We present an extended version of the results on CIFAR-100, summarized in Sec.~$6$ in the main paper. Specifically, for each experiment (Sections~$6.1$, $6.2$, and $6.3$), we provide additional figures to complete the information shown in Tables~$1$ and $2$ in the main paper. We also present an extended version of Fig.~$3$ from the original paper.

\subsection*{Fixed memory size}\label{sec:results-sup}
\begin{figure}[h]
    \centering
    \includegraphics[width = \textwidth]{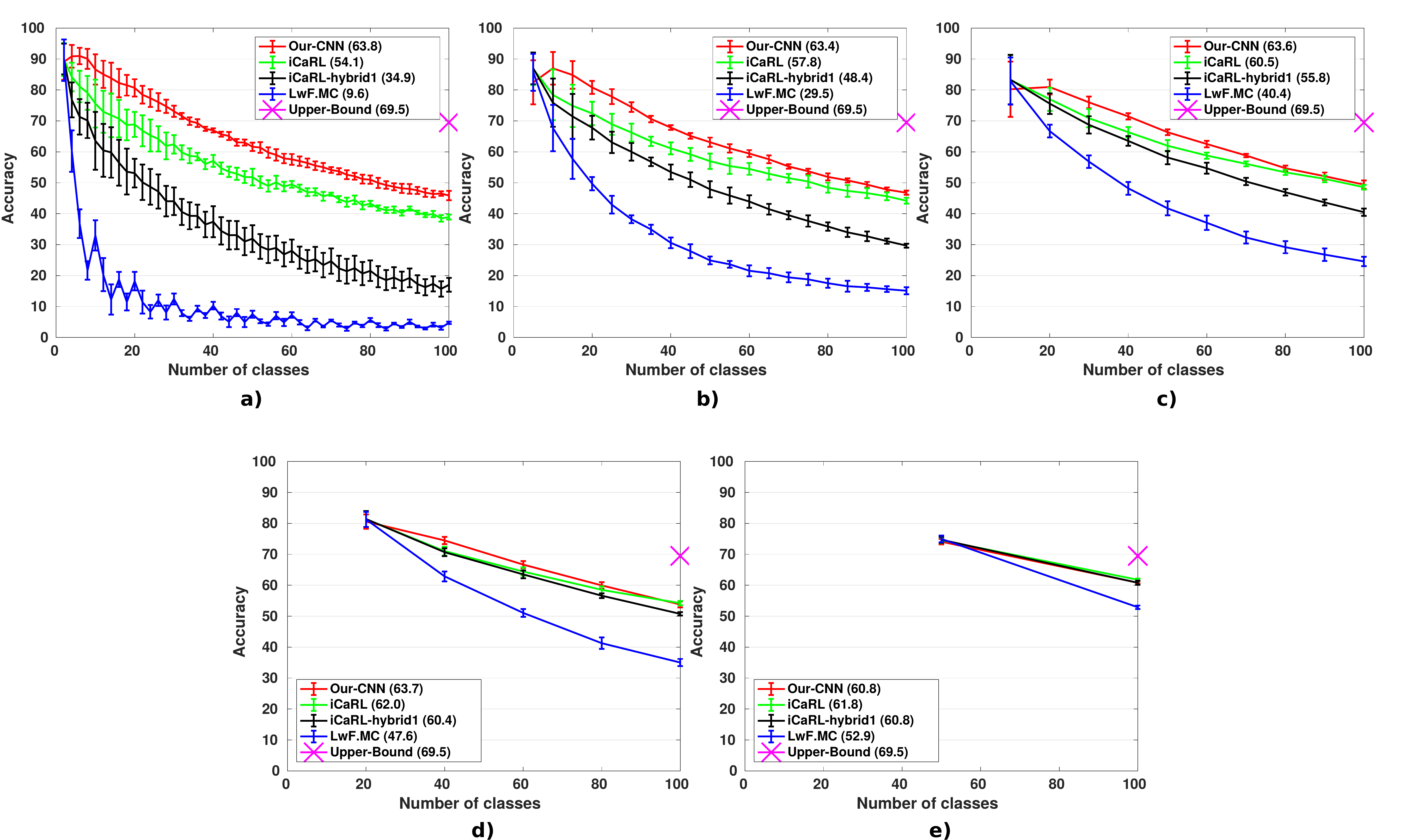}
    \caption{\textbf{Accuracy on CIFAR-100 with fixed memory size.} Average and standard deviation of 5 runs with {(a)} 2, {(b)} 5, {(c)} 10, {(d)} 20, and {(e)} 50 classes per incremental step. The average over all the incremental steps is shown in parentheses for each method. (Best viewed in pdf.)}
    \afterCaption
    \label{fig:results-sup}
\end{figure}

We present Fig.~\ref{fig:results-sup} in this appendix, which is an extended version of Fig.~$3$ in the main paper. It shows the accuracy of all the incremental steps. As commented in the main paper (Tab.~$1$(a) and Fig.~$3$), our approach shows the best performance for incremental steps with $2$, $5$, $10$, and $20$ classes. This improvement is specially remarkable for steps with small number of classes (i.e., 2 and 5). In the other cases, the results are similar to iCaRL in the final incremental steps. In comparison to the iCaRL variant hybrid1, using a CNN classifier, we show a significant improvement in nearly all the cases. For incremental steps with $50$ classes, the performance of our approach is similar to hybrid1, and slightly lower than iCaRL. We believe this is due to the unbalanced number of samples from the new and old classes.

\subsection*{Fixed number of samples} \label{sec:incremental_analysis-sup}
\begin{figure}[h]
    \centering
    \includegraphics[width=1.0\textwidth]{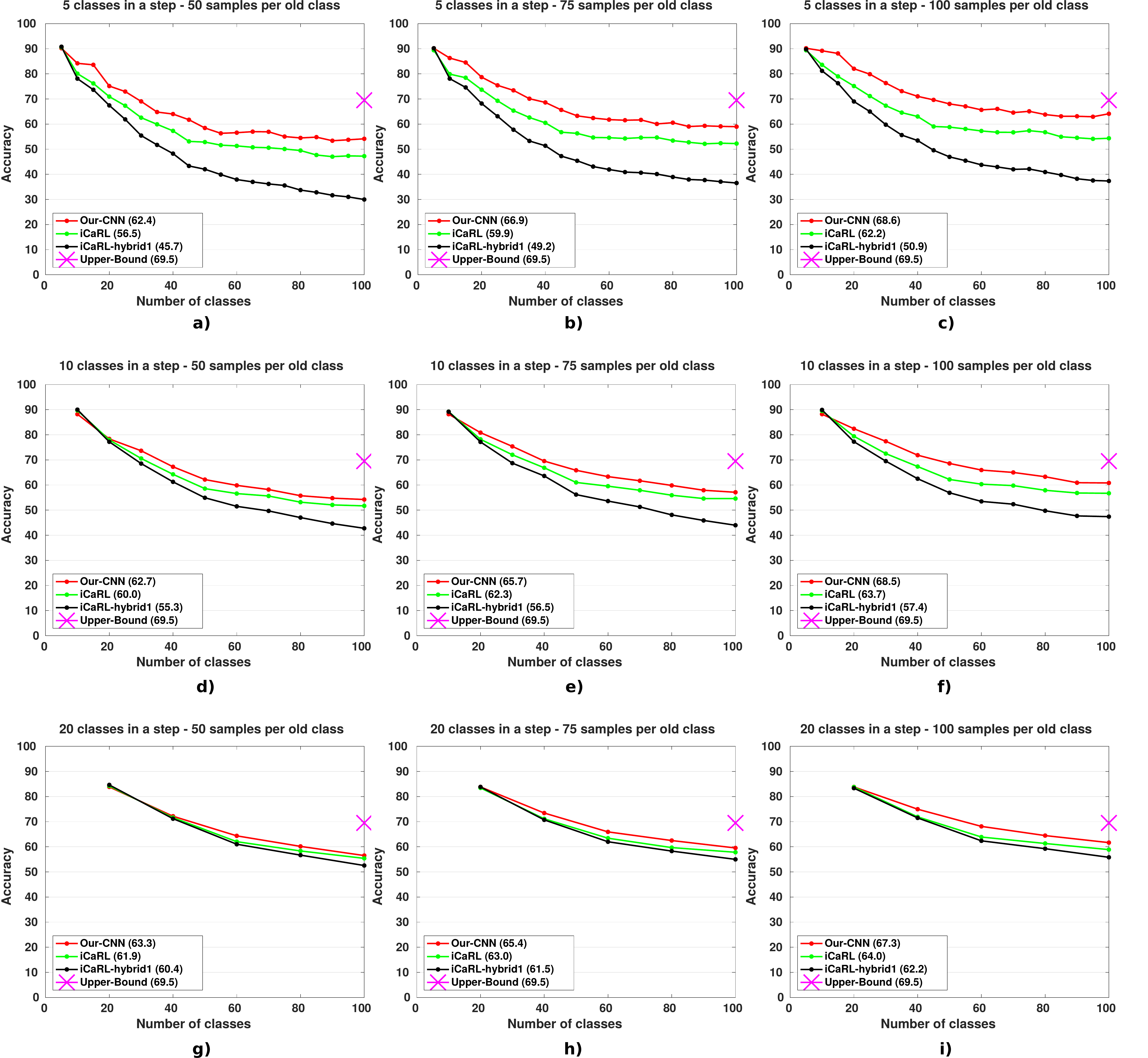}
    \caption{\textbf{Accuracy on CIFAR-100 with fixed number of samples.} Results for 5 (a, b, c), 10 (d, e, f) and 20 (g, h, i) classes with 50, 75, and 100 samples for each of the old classes respectively. The average over all the incremental steps is shown in parentheses for each method. (Best viewed in pdf.)}
    \afterCaption
    \label{fig:results-sup-images}
\end{figure}

We provide a new illustration, Fig.~\ref{fig:results-sup-images} that complements Tab.\ 2(a) in the main paper. It shows the accuracy of incremental steps with $5$, $10$, and $20$ classes, using $50$, $75$, and $100$ exemplars for each of the old classes. Our approach shows the best results, and the improvement is larger in the challenging cases with fewer number of classes per incremental step. Note in particular, the performance of our approach with $5$ classes and $100$ samples for each old class, i.e., Fig.~\ref{fig:results-sup-images}(c), with only $20\%$ of the original data. Here, the performance also approaches the upper bound (shown with a cross in the figure).

\clearpage
\section*{Appendix B: Comparison with Gradient Episodic Memory}
The approach presented in~\cite{lopez2017nips} for continual learning is based on an episodic memory, which stores a subset of samples from each task trained. During the continual or incremental learning, this memory stores new samples from the previously trained tasks. To learn new classes and, at the same time, to retain the knowledge from the previously trained classes, the loss functions are used as inequality constraints. Thus, the losses can decrease due to an improvement in  the solution but they cannot increase, avoiding worse solutions and loss of previous knowledge. 

We compare~\cite{lopez2017nips} with our method using the code provided%
\footnote{\url{https://github.com/facebookresearch/GradientEpisodicMemory}} under two settings. In the first one, proposed in~\cite{lopez2017nips}, the task of each test sample is known a priori, and only the scores from this task are used at test time. In the second setting, the task is unknown and scores from all the tasks are taken into account during the test phase. Better accuracies are expected in the first experiment, as only scores for the classes belonging to the correct task are evaluated.

\subsection*{A priori known task}
We follow the setup proposed in~\cite{lopez2017nips}. The $100$ classes in CIFAR-100 are split into groups of 5 classes, and trained in an incremental way. At test time, for an input sample, the group of classes (or task) is known a priori, so the output probabilities are truncated to these classes. This is a simplified version of the problem as the output class is selected among the five classes, instead of all the trained classes. Tab.~\ref{tab:GEM} shows the final accuracy on CIFAR-100 with different memory sizes. This final accuracy is obtained at the end of each incremental learning step by averaging the individual accuracy of each task like in~\cite{lopez2017nips}.  The class order is identical for all the evaluated methods, to ensure that the results are comparable. As seen in the table, our approach performs better in all cases, specially when the memory size is small.

\begin{table}[ht]
\centering
\begin{tabular}[t]{l|c|c|c|c}
\hline 
memory size & 200 & 1280 & 2560 & 5120 \\ 
\hline \hline
GEM & 48.0 & 62.6 & 64.6 & 67.8 \\
Ours & \textbf{77.9} & \textbf{86.1} & \textbf{88.4} & \textbf{90.7} \\
\hline 
\end{tabular}
\caption{Accuracy for different memory sizes on CIFAR-100.}
\afterCaption
\afterCaption
\afterCaption
\afterCaption
\afterCaption
\label{tab:GEM}
\end{table}

\subsection*{Unknown task}
\begin{figure*}[t]
    \centering
    \includegraphics[width=0.6\textwidth]{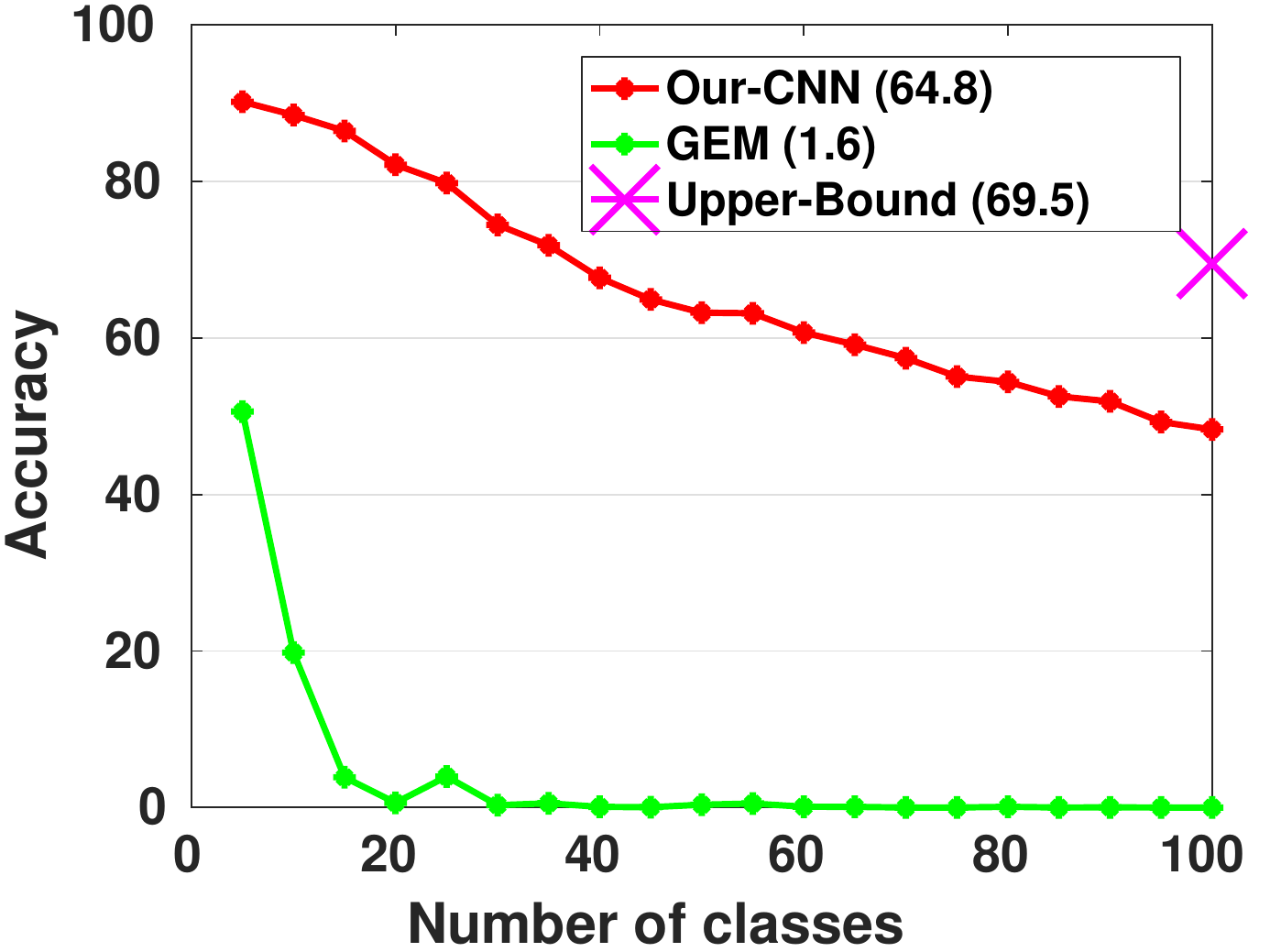}
    \caption{\textbf{Accuracy on CIFAR-100.} Execution with 5 classes per incremental step. Average of the incremental steps is shown in parentheses for each method. (Best viewed in pdf.)}
    \afterCaption
    \label{fig:results-gem}
\end{figure*}

We use the standard incremental learning setup from~\cite{rebuffi2017cvpr}. Again, the $100$ classes in CIFAR-100 are split into groups of 5 classes and trained in an incremental way. However, during test time, the task index is unknown and the method must differentiate among classes from different tasks. Fig.~\ref{fig:results-gem} shows the incremental accuracy per step on CIFAR-100, with a memory size of 2000 samples. This incremental accuracy is obtained at the end of the incremental step on the test samples from the trained classes (old and new). The `Upper-Bound' result, shown in Fig.~\ref{fig:results-gem} with a cross (in magenta) in the last step, is obtained by training a non-incremental model with all the training samples.

Our method shows a good performance without knowing the task of each test sample. The accuracy decreases with the number of classes as it has to retain more knowledge from previous classes, and the number of samples from those classes decreases continuously due to a fixed total number of samples. The accuracy of GEM decreases dramatically with the number of classes. Thus, in scenarios where the task is unknown a priori, GEM fails to perform competitively.

\clearpage
\setcounter{figure}{0}
\section*{Appendix C: Comparison with similar classes}
We now test our hypothesis that using an external classifier (like NCM in iCaRL~\cite{rebuffi2017cvpr}), instead of an end-to-end approach, should produce lower results when the dataset contains similar classes. For example, in a face recognition problem trained in an incremental way, all classes have a similar mean as only small details of the face change. Therefore, an end-to-end classifier trained together with the feature extractor should obtain better results, as it has learned to deal with the small differences among classes.

We perform an experiment with a subset of 10 vehicle classes in CIFAR-100. We use the standard training/test sets with 5 incremental steps of 2 classes each. For ImageNet, we use a subset of 120 dog breeds with 12 incremental steps of 10 classes each. For the cars experiment we use a memory size of 200 samples, and for dog breeds we use 2400 samples.

\begin{figure*}[hb]
    \centering
    \includegraphics[width=1.0\textwidth]{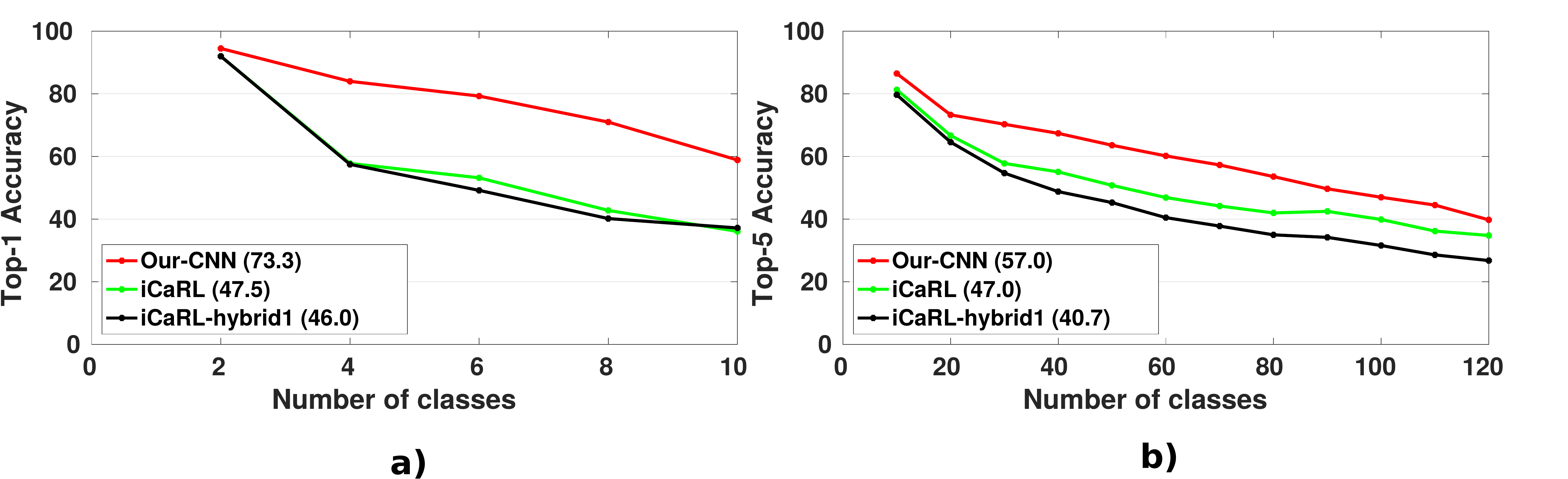}
    \caption{\textbf{Accuracy on subsets with similar classes.} (a) Two classes per incremental step on 10 types of vehicles in CIFAR-100. (b) Ten classes per incremental step on 120 dog breeds in ImageNet. Average of the incremental steps is shown in parentheses for each method. (Best viewed in pdf.)}
    \afterCaption
    \label{fig:results-others}
\end{figure*}

Fig.~\ref{fig:results-others}(a) and Fig.~\ref{fig:results-others}(b) show the results of the vehicles/cars and the dog breeds experiments respectively. In both cases, our end-to-end approach achieves the best results with a significant boost in average accuracy over other methods. For the cars experiment, our model obtains an average accuracy of $73.3\%$ while iCaRL is $47.5\%$. In this case, hybrid1 shows similar results to iCaRL. Similar behaviour is observed in the case of the dog breeds experiment, but with a slightly lower boost in performance.

\end{document}